\documentclass[draft ,onecolumn]{IEEEtran}
\ifCLASSINFOpdf
\else
\fi

\IEEEoverridecommandlockouts
\usepackage{cite}
\usepackage{amsmath,amssymb,amsfonts}
\usepackage{amssymb}
\usepackage{breqn}
\usepackage{cleveref}
\usepackage[lined,ruled,linesnumbered,commentsnumbered]{algorithm2e}
\usepackage{booktabs}
\usepackage{cases}
\usepackage{graphicx}
\usepackage{multirow}
\usepackage{mathrsfs}
\usepackage{textcomp}
\usepackage{xcolor}
\usepackage{subfigure}

\newtheorem{myDef}{Definition}

\def\BibTeX{{\rm B\kern-.05em{\sc i\kern-.025em b}\kern-.08em
    T\kern-.1667em\lower.7ex\hbox{E}\kern-.125emX}}

\begin{document}



\title{Evolutionary Dynamic Multi-objective Optimization Via Regression Transfer Learning}
\bibliographystyle{IEEEtran}

\author{Zhenzhong~WANG,
	Min JIANG$^*$,
	Xing GAO$^*$,
	Liang FENG,
	Weizhen HU,
	and Kay Chen TAN
	\IEEEcompsocitemizethanks{\IEEEcompsocthanksitem Z.WANG and M. JIANG and X.GAO and W.HU are with the school of informatics, Xiamen University, China, Fujian, 361005. Min JIANG and Xing GAO are corresponding authors and email: minjiang@xmu.edu.cn, gaoxing@xmu.edu.cn.

	\IEEEcompsocthanksitem L.FENG is with the College of Computer Science, Chongqing University, Chongqing, China 400044.

		\IEEEcompsocthanksitem KC TAN is with the Department of Computer Science, City University of Hong Kong.}
	
}










\maketitle

\begin{abstract}

Dynamic multi-objective optimization problems (DMOPs) remain a challenge to be settled, because of conflicting objective functions change over time. In recent years, transfer learning has been proven to be a kind of effective approach in solving DMOPs. In this paper, a novel transfer learning based dynamic multi-objective optimization algorithm (DMOA) is proposed called regression transfer learning prediction based DMOA (RTLP-DMOA). The algorithm aims to generate an excellent initial population to accelerate the evolutionary process and improve the evolutionary performance in solving DMOPs. When an environmental change is detected, a regression transfer learning prediction model is constructed by reusing the historical population, which can predict objective values. Then, with the assistance of this prediction model, some high-quality solutions with better predicted objective values are selected as the initial population, which can improve the performance of the evolutionary process. We compare the proposed algorithm with three state-of-the-art algorithms on benchmark functions. Experimental results indicate that the proposed algorithm can significantly enhance the performance of static multi-objective optimization algorithms and is competitive in convergence and diversity.

\end{abstract}


\section{INTRODUCTION}

Many optimization problems in the real world\cite{6358934}\cite{Raquel_2013} involve multiple optimization functions which conflict with each other and change over time. These dynamic optimization problems are called Dynamic Multi-objective Optimization Problems (DMOPs)\cite{Yang2017Multi}. For example, in the design of job scheduling systems\cite{8359103}, a number of decision variables, such as procedures, components, and operation time, are involved, which determine objective functions of energy consumption, production, and stability. These conflicting objective functions always change with time. Hence, efficient DMOAs should rapidly arrange scheduling schemes according to the changing environments, and this ability is critical to robust scheduling systems.

In recent years, in order to solving DMOPs, a variety of DMOAs have been proposed. These existing methods can be roughly grouped into the following three categories: The first category of DMOAs is based on maintaining diversity. Gong \textit{et al.} \cite{8694000} proposed a general framework to decompose decision variables into two subpopulations according to the interval similarity between each decision variable and interval parameters, and a strategy on the basis of change intensity is adopted to track the POF. In \cite{JIANG2018203}, Jiang \textit{et al.} developed a framework based on domain adaptive and non-parametric estimation to keep the exploration-exploitation of DMOPs in terms of temporal and spatial views. The second is a memory-based method. Chen \textit{et al.} \cite{7886303} implemented a dynamic two-archive strategy to simultaneously maintain two co-evolving populations. One population is concerned on convergence while the other focuses on diversity. Branke \textit{et al.} \cite{Branke} proposed a memory scheme to enhance the evolutionary process. In this algorithm, some excellent solutions are saved which can be used for guiding towards to optimal solutions. The third category of DMOAs is based on prediction. Muruganantham \textit{et al.} \cite{MurugananthamTV16} presented a population prediction strategy based on the Kalman filter technique. The Kalman filter technique\cite{Welch2001Kalman} can guide the search for new Pareto-optimal solutions to generate a large number of high-quality initial individuals. Then, the algorithm finds the optimal at this moment based on a decomposition-based differential evolution algorithm. Rong \textit{et al.} \cite{8388721} presented a prediction model to track the moving POS by clustering the whole population into several subpopulations. In addition, the number of clusters depends on the intensity of environmental change. Zhou \textit{et al.} \cite{zhou2014population} proposed a population prediction method to predict a whole population instead of predicting some isolated points. The algorithm uses center points to predict the next center point, and the previous manifolds are used to estimate the next manifold. The optimal population at this moment is determined based on a decomposition-based differential evolution algorithm. Hu \textit{et al.} \cite{8790005} designed a promising approach based on Incremental Support Vector Machine (ISVM)\cite{6876186} classifier in solving DMOPs, the ISVM is trained from the past Pareto-optimal set, then high-quality initial individuals are filtered through the trained ISVM. Jiang \textit{et al.} \cite{8100935} presented a framework based on transfer learning \cite{7349204} to predict an effective initial population for solving DMOPs. The transfer component analysis (TCA)\cite{7349204} is used in this framework for the domain adaptation problem\cite{LU201514}.


Traditional machine learning approaches are usually based on the assumption that the samples follow the Independent Identically Distributed (IID). Nevertheless, this hypothesis will be broken when dealing with DMOPs, since the solution distribution fails to satisfy the IID hypothesis. Although there is a DMOA based on transfer learning. However, it leads to poor diversity when samples clustering in the high dimensional latent space created by TCA.

In this paper, a regression transfer learning prediction based DMOA (RTLP-DMOA) is proposed. The algorithm aims to generate an excellent initial population to enhance the ability of existing multi-objective optimization algorithms for DMOPs. When the environment has changed, a regression transfer learning prediction model is constructed by utilizing the historical population information which can predict objective values in the new environment. Then, with the assistance of this regression prediction model, some high-quality solutions with better predicted objective values can be identified and selected as an initial population, which can improve the individuals' performance of the evolutionary process significantly.

The contributions of this work are as follows: 1) The proposed algorithm can make full use of historical information and predict high-quality initial population to improve the evolutionary performance of the existing static multi-objective optimization algorithms (SMOAs) in solving DMOPs. 2) The proposed algorithm can overcome the difficulty that solution distributions fail to meet the IID hypothesis. Compared with other prediction methods, the RTLP-DMOA is promising.


The rest of the paper is organized as follows: In Section II, we describes the basic concepts of DMOPs and presents the related transfer learning method used in the RTLP-DMOA. Section III gives the designed RTLP-DMOA in detail. In Section IV, experimental results and analysis are shown. Conclusions are drawn in Section V.
 
\section{PRELIMINARY STUDIES}
\subsection{Dynamic Multi-objective Optimization}
The mathematical form of DMOPs is as follows:

\begin{equation}
min \ F(x,t)=<f_1(x,t),f_2(x,t),...,f_m(x,t)>  
\end{equation}
where $x \in \Omega$, and $x=<x_1,x_2,...,x_n>$ is the $n$-dimensional decision vector, and $t$ is the environment variable. $F=<f_1,f_2,...,f_m>$ is the $m$-dimensional objective vector. The goal of DMOAs is to find solutions at environment $t$ so that all objectives are as small as possible. Nevertheless, one solution cannot satisfy the minimum of all conflicting objectives. Hence, a trade-off method called Pareto dominance is introduced to compare these solutions. The set of optimal trade-off solutions is called the Pareto-optimal solutions (POS) in the decision space and the Pareto-optimal front (POF) in the objective space\cite{deb2001multi}.
\begin{myDef}
	\textbf{(Dynamic Decision Vector Domination)} At environment $t$, a decision vector $x_1$ Pareto-dominates another vector $x_2$ denoted by $x_1\succ_t x_2$, if and only if
	\begin{equation}\label{chap2:equ2}
	\begin{cases}
	f_j(x_2,t)& \ge f_j(x_1,t),\quad \forall j=1,...,m\\
	f_j(x_2,t)& > f_j(x_1,t),\quad \exists j=1,...,m
	\end{cases}
	\end{equation}
	
\end{myDef}

\begin{myDef}
	\textbf{(Dynamic Pareto-Optimal Set, DPOS)} If a decision vector $x^*$ at environment $t$ satisfies
	\begin{equation}
	DPOS=\left\{x^*|\nexists x, x\succ_t x^*\right\},
	\end{equation}
	then all $x^*$ are called dynamic Pareto-optimal solutions, and the set of dynamic Pareto-optimal solutions is called the dynamic POS (DPOS).
\end{myDef}

\begin{myDef}
	\textbf{(Dynamic Pareto-Optimal Front, DPOF)} DPOF is the Pareto-optimal front of the DPOS for the DMOPs at the environment $t$
	\begin{equation*}
	DPOF=\{y^*|y^*=F(x^*, t),x^*\in DPOS\}.
	\end{equation*}
\end{myDef}

\subsection{TrAdaboost.R2}

TrAdaboost \cite{Dai2007Boosting} is a classification algorithm based on the boosting method. The aim of TrAdaBoost is to filter out dissimilar samples in the past source domain to those in the target domain. In this way, TrAdaboost improves the classification accuracy. The source data set is combined with the target domain set to form a single data set. At each boosting step, TrAdaBoost increases the relative weights of target instances that are misclassified. When a source instance is misclassified, however, its weight is decreased. In this way, TrAdaBoost makes use of those source instances that are most similar to the target data while ignoring those that are dissimilar. In \cite{Pardoe2010Boosting}, the authors introduce TrAdaboost-based algorithms for transfer regression task, called TrAdaboost.R2.

TrAdaboost.R2 is an ensemble method in which each weak regression hypothesis $h_i$ ($i=1,...,K$) can map the source domain data set $X_{source}$ and the target domain data $X_{target}$ to $Y\in R$. A strong regression hypothesis $h$ is determined by combining these weak hypotheses. In each training round, TrAdaboost.R2 increases the relative weights of instances from the target domain. Meanwhile, TrAdaboost.R2 decreases the weights of the instances from the source domain. When the regression error of a instance caused by $h_i$ is large, $h_i$ has a substantial influence on the changing weight of the instance. In this way, TrAdaboost.R2 reuses source instances that are most similar to the target data and ignores those that are dissimilar. In the next round, these modified weights are inputed into the next regression hypothesis $h_{i+1}$, instances that are dissimilar to the target domain weaken their impacts of learning process, and instances with large weights help the learning algorithm in training better regressions.

\section{PROPOSED ALGORITHM}

The framework of RTLP-DMOA is illustrated in Algorithm \ref{alg:dmoea}. In brief, RTLP-DMOA initializes randomly a population $initPop$ with size $N$, and then executes a SMOA to optimize the population at environment $t=0$. If environmental changes are detected, the environment variable $t$ is updated as $t=t+1$. Then, the last population $P_{t-1}$ is inputted into the procedure of regression transfer. In the procedure of regression transfer, a regression hypothesis $h^t$ is determined with historical information which can predict objective vectors of individuals at the new environment. Next, in the procedure called initial population prediction, the $h^t$ is employed to predict the objective vectors and some high-quality individuals are selected according to their predictive objective vectors. These individuals are regarded as an excellent initial population $initPop$ and inputted into a SMOA to accelerate the evolutionary process. The details of RTLP-DMOA are presented in the following section.

\begin{algorithm}
  \caption{RTLP-DMOA}
  \label{alg:dmoea}
  \KwIn{The dynamic multi-optimization problem $F_t(\cdot)$, the population size $N$, a $SMOA$}
  \KwOut{the population $P_t$ of $F_t(\cdot)$}
  Initialize the environment variable $t=0$\;
  Initialize randomly a population $initPop$ with $N$\;
  $P_t=\textbf{SMOA}(initPop,F_t)$\;
  \While{the environment has changed}{
  $t=t+1$\;
  $h^t=\textbf{Regression-Transfer}(F_t, F_{t-1}, P_{t-1})$\;

  $initPop=\textbf{Initial-Population-Prediction}(h^t,N)$\;
  $P_t=\textbf{SMOA}(initPop,F_t)$\;
  \Return $P_t$
  }
\end{algorithm}

\subsection{Regression Transfer}

The regression transfer process returns a strong regression hypothesis $h^t$ for environment $t$. The strong regression hypothesis $h^t$ adapts to the solution distribution at current environment. When an individual $x$ is given, $h^t(x)$ outputs a predicted objective vector of $x$. Therefore, in the subsequent process of RTLP-DMOA, an excellent individual $x$ with better predicted objective vectors $h^t(x)$ can be selected as a member of the initial population.

The strong regression hypothesis $h^t$ is integrated with several weak regression hypotheses $h^t_i$ ($i=1,...,K$, $K$ is the maximum number of iterations for training). These weak regression hypotheses are trained with the past population information. The last population $P_{t-1}$ combined with their objective values $F_{t-1}(P_{t-1})$ are regarded as source domain set $D_{source}$. The target domain set $D_{target}$ is comprised of $\hat{P_{t}}$ which is sampled from $U(a, b)$ in the current decision space and their objective values $F_t(\hat{P_{t}})$, where $a$ and $b$ are the lower bound and upper bound of the decision variable at environment $t$. $D_{source}$ and $D_{target}$ are combined into a set $D$ as the training data. 

The process for training weak regression hypotheses is as follows: First of all, the weight vector $w^1_i(x)$ is initialized as $\frac{1}{|D|}$, $w^t_i(x)$ denotes the weight of $x$ for training $h_i^t$ at environment $t$. In the main training loop, for training $h_i^t$, a Support Vector Regression (SVR)\cite{basak2007support} is implemented as a basic learner to obtain the weak regression hypothesis $h^t_i$ from $D$ and $w_i^t$. Then, the adjusted error $e_i^t(x)$ of each individual $x\in D$ for training $h_i^t$ is calculated as
\begin{equation}\label{eq:aei2}
       e_i^t(x)=
       \begin{cases}
        \frac{|F_{t-1}(x)-h_i^t(x)|}{E_i^t}, x\in D_{source}\\
        \frac{|F_t(x)-h_i^t(x)|}{E_i^t}, x\in D_{target}
      \end{cases},
\end{equation}
where $E_i^t$ is the maximum error, it is described as
\begin{equation}\label{eq:me}
\begin{split}
       E_i^t=\max \{ \max \{ F_{t-1}(x)-h_i^t(x)| x\in D_{source}\},\\ \max\{ F_t(x)-h_i^t(x)| x\in D_{target}\} \}.
       \end{split}
\end{equation}
The $e_i^t(x)$ is bigger when the difference between the predicted object vector $h^t_i(x)$ and the true objective vector $F(x)$ become bigger, and the adjusted error $\epsilon_i^t$ for $h^t_i$ is calculated as
\begin{equation}\label{eq:aeh}
       \epsilon_i^t=\sum_{x\in D_{target}}e_i^t(x) w^t_i(x).
    \end{equation}
When $e_i^t$ is small, $\epsilon_i^t$ becomes smaller. Next, the weight vector is updated according to $e_i^t$ and $\epsilon_i^t$: If a training individual from the $D_{source}$ has a bigger $e_i^t$, the individual may be more dissimilar to the distribution of the target domain. Therefore, its training weight must be reduced more. However, if a training individual from the target domain has a bigger $e_i^t$, then its training weight should be increased more for $h_i^t$ to adapt the target domain. So, the weights can be updated as
\begin{equation}\label{eq:aw}
       w_{i+1}^{t}(x)=
      \begin{cases}
        w_i^t(x)\beta^{e_i^t(x)}, x\in D_{source}\\
        w_i^t(x)\beta_i^{-e_i^t(x)}, x\in D_{target}
      \end{cases},
    \end{equation}
where $\beta_i=\epsilon_i^t/(1-\epsilon_i^t)$, and $\beta=1/(1+\sqrt{2\ln |D_{source}|/K})$. In this way, individuals adapted to the solution distribution of the target domain have large weights; otherwise, they have small weights. Then, modified weights $w_{i+1}^{t}$ are inputted into next SVR to learn $h^t_{i+1}$. Thus, in the next round, individuals with low weights that are dissimilar to the target domain weaken their impacts of the learning process and those with large weights will help the learning algorithm train better regression hypotheses. These weak regression hypotheses ($h^t_i$, $i=1,...,K$) may gradually adapt to the target domain. After $K$ iterations, we obtain the final $\left \lceil \frac{K}{2} \right \rceil$ weak regression hypotheses and combine them to acquire a strong regression $h^t$. 

The details of regression transfer are shown in Procedure {\sc Regression Transfer}.

\begin{procedure}
  \caption{(\sc Regression Transfer)}
  \label{alg:rtm}
  \KwIn{Dynamic optimization functions $F_t$ and $F_{t-1}$, the population $P_{t-1}$}
  \KwOut{A regression hypothesis $h^t$}
  $D_{source}=\{P_{t-1},F_{t-1}(P_{t-1})\}$\;
  Sample $\hat{P_{t}}$ from $U(a, b)$\;
  $D_{target}=\{\hat{P_{t}}, F_t(\hat{P_{t}})\}$\;
  $D=D_{source}\cup D_{target}$\;
  Initialize the weight vector $w^1_i=\frac{1}{|D|}$\;
  Set the maximum number of iterations $K$\;
  \For{$i=1$,\ldots,$K$}{
  Call a SVR with $D$ and $w^t_i$, and get a regression hypothesis $h^t_i:X\rightarrow R^m $\;
  Calculate the adjusted error according to (\ref{eq:aei2}) for each individuals\;
  Calculate the adjusted error of $h^t_i$ according to (\ref{eq:aeh})\;
  Update the weight vector according to (\ref{eq:aw})\;
  }
  \Return 
      $h^t=$ the weighted median of $h^t_i(x)$ for the final $\left \lceil \frac{K}{2} \right \rceil$, and using $\ln(1/\beta_i)$ as the weight for hypothesis $h^t_i(x)$.
\end{procedure}

\subsection{Initial Population Prediction}

\begin{procedure}
  \caption{(\sc Initial Population Prediction)}
  \label{alg:ipp}
  \KwIn{A regression hypothesis $h^t$, the population size $N$}
  \KwOut{An initial population $initPop$}
  Sample $P_{t}^{test}$ from $U(a, b)$\;
  $initPop=\varnothing$\;
  Use $h^t$ to predict the objective value $h^t(P_{t}^{test})$\;
  $\mathcal{F}=\textbf{fast-non-dominated-sort}(P_{t}^{test},h^t(P_{t}^{test}))$\;
  $i=1$\;
  \While{$|initPop|+|\mathcal{F}_i|\leq N$}{
  $initPop=initPop\cup \mathcal{F}_i$; //$\mathcal{F}_i$ represents the $i$-th non-dominated front.\\
  $i=i+1$\;
  }
  Add Gaussian noise to $initPop$ until the population size is $N$\;
  \Return $initPop$
\end{procedure}

In this section, the initial population prediction is utilized to identify some excellent solutions as the initial population with the assistance of $h^t$. 

To begin with, a test population $P_{t}^{test}$ is sampled from $U(a, b)$, where $a$ and $b$ are the lower bound and upper bound of the decision variable at environment $t$. Then, objective values $h^t(P_{t}^{test})$ are predicted, and the non-dominated front $\mathcal{F}$ can be determined by fast non-dominated sort\cite{996017} according to predicted objective values. Then, we select the first $i$ non-dominated fronts as $initPop$ and limit the size of $initPop$ does not exceed the population size $N$. Next, some Gaussian noises are added to $initPop$ until the population size is $N$. The initial population $initPop$ is to accelerate the evolutionary process and improve the evolutionary performance for the current environment. 

The details of initial population prediction are presented in Procedure {\sc Initial Population Prediction}.

\section{EXPERIMENTS}

\subsection{Compared Algorithms}

Three DMOAs used for comparison in the experiment are as follows:
\begin{enumerate}
  
  \item dCOEA\cite{goh2009competitive}: It is a DMOA with dynamic competition-cooperative co-evolution.
  \item PPS\cite{Aimin_Zhou_2014}: It is a DMOA based on population prediction.
  \item SGEA\cite{jiang2017steady}:  It is a DMOA based on steady state and maintaining population diversity.
\end{enumerate}

\subsection{Test Problems}

All compared algorithms are evaluated on 8 benchmark DMOPs selected from FDA\cite{farina2004dynamic2222} and DMOP\cite{goh2009competitive}. The FDA benchmark comprises FDA1, FDA2, FDA3, FDA4, and FDA5. The DMOP benchmark contains dMOP1, dMOP2, and dMOP3. 

DMOPs is divided into three categories: Type I problem indicates POS changes, but the POF does not change. Type II problem indicates changes in POS and POF. Type III problem implies the POF changes but the POS does not change. 

FDA1, FDA4, and dMOP3 belong to Type I problem. FDA3, FDA5, and dMOP2 belong to Type II problem. Type III contains FDA2 and dMOP1. 

The dynamics of a DMOP is controlled by 
\begin{equation}\label{equ:MS}
    t=\frac{1}{n_t}\left \lfloor \frac{\tau}{\tau_t}  \right \rfloor,
\end{equation}
where $\tau$, $n_t$, and $\tau_t$ refer to the generation counter, severity of change, and frequency of change, respectively.

\subsection{Performance Indicators}


1) The Inverted Generational Distance (IGD) metric \cite{Zhang2008RM} can measure the convergence of obtained solutions. A small IGD value represents the convergence of the solution is improved. IGD is defined as

  \begin{equation}\label{chap2:equ2}
      IGD(POF^{*},POF)=\frac{1}{N}\sum_{p^{*}\in POF^{*}}\min_{p\in POF}{\left \| p^*-p \right \|}^2,
  \end{equation}
where $POF^{*}$ is the true POF of a multi-objective optimization problem, and $POF$ is an approximation set of POF obtained by a multi-objective optimization algorithm and $N$ is the number of individuals in the $POF^{*}$.

The MIGD\cite{7886303} metric is a variant of IGD. The MIGD can be described as the average of the IGD values in all environments during a run.
  \begin{equation}\label{chap2:equ2}
      MIGD(POF^*_t,POF_t)=\frac{1}{|T|}\sum_{t\in T}IGD(POF^*_t,POF_t),
  \end{equation}
  where $T$ is a set of discrete time points during a run and $|T|$ is the cardinality of $T$.

2) The Maximum Spread (MS)\cite{Aimin_Zhou_2014} can quantify the extent of obtained solutions covers the true POF. A large MS value indicates additional coverage for the true POF by solutions obtained by the algorithm. MS is calculated as follows:
  \begin{equation}\label{equ:MS}
    MS=\sqrt{\frac{1}{m}\sum_{k=1}^{m} \left[ \frac{ \min \left[ F_k^{max},f_k^{max} \right]-\max \left[F_k^{min},f_k^{min} \right]}{F_k^{max}-F_k^{min}} \right]^2},
  \end{equation}
where $F_k^{max}$ and $F_k^{min}$ represents maximum and minimum of $k$-th objective in true POF, respectively; and $f_k^{max}$ and $f_k^{min}$ represent the maximum and minimum of $k$-th objective in the obtained POF, respectively. This metric is also modified for evaluating DMOAs.

\subsection{Parameter Settings}

Parameter settings in RTLP-DMOA are as follows: We set the size of the population $N$ to 100 and set the number of iterations $K$ for training $h^t$ to 10. The size of $\hat{P_{t}}$ and $P_{t}^{test}$ are set to 50 and 500, respectively. We choose RM-MEDA\cite{Zhang2008RM} as the SMOA optimizer for RTLP-DMOA, and the number of cluster is 4 in RM-MEDA. The parameters in SVR are set by default\cite{Chang:2011:LLS:1961189.1961199}.

Consistent with the experimental configuration in this study \cite{jiang2017steady}: We fix the $n_t$ to 10. The frequency of change $\tau_t$ values are 5, and 10. The number of iterations of compared algorithms is $3\times n_t\times \tau_t+50$, of which 50 are the number of iterations at the initial time. Hence, in each population of configurations, the problem is changed by $3\times n_t$ times.

\subsection{Experimental Results}

Experimental comparison results of RTLP-DMOA with other three state of the art DMOAs. MIGD values and MS values are presented in Tables \ref{tab:igd} and Table \ref{tab:ms}, respectively. The best metric values are highlighted in bold.

\begin{table*}[htbp]
  \centering
  \caption{MEAN AND STANDARD DEVIATION VALUES OF MIGD METRIC FOR DIFFERENT DYNAMIC TEST SETTINGS}
    \begin{tabular}{cccccc}
    \toprule
    Problem & {$\tau_t$,$n_t$}     & RTLP-RM-MEDA & dCOEA & PPS   & SGEA \\
    \midrule
    \multirow{2}[2]{*}{FDA1} & (5,10) & \textbf{0.0051(0.0013)} & 0.0661(0.0128) & 0.2061(0.0769) & 0.0338(0.0081) \\
          & (10,10) & \textbf{0.0049(0.0011)} & 0.0413(0.0068) & 0.0476(0.0204) & 0.0132(0.0025) \\
    \midrule
    \multirow{2}[2]{*}{FDA2} & (5,10) & \textbf{0.0228(0.0046)} & 0.0774(0.0390) & 0.0888(0.0348) & 0.0121(0.0014) \\
          & (10,10) & 0.0223(0.0529) & 0.0491(0.0329) & 0.0619(0.0107) & \textbf{0.0083(0.0006)} \\
    \midrule
    \multirow{2}[2]{*}{FDA3} & (5,10) & 0.1425(0.0066) & 0.2640(0.0355) & 0.4143(0.0101) & \textbf{0.0612(0.0327)} \\
          & (10,10) & 0.1455(0.0081) & 0.1910(0.0338) & 0.2003(0.0183) & \textbf{0.0405(0.0180)} \\
    \midrule
    \multirow{2}[2]{*}{FDA4} & (5,10) & \textbf{0.1116(0.0092)} & 0.1604(0.0066) & 0.3191(0.0203) & 0.1603(0.0642) \\
          & (10,10) & \textbf{0.1189(0.0091)} & 0.1296(0.0048) & 0.2196(0.0215) & 0.1241(0.0664) \\
    \midrule
    \multirow{2}[2]{*}{FDA5} & (5,10) & \textbf{0.3615(0.0027)} & 0.4387(0.0469) & 0.6577(0.0318) & 0.5221(0.0395) \\
          & (10,10) & \textbf{0.3612(0.0053)} & 0.3691(0.0403) & 0.5037(0.0355) & 0.4002(0.0088) \\
    \midrule
    \multirow{2}[2]{*}{DMOP1} & (5,10) & 0.0469(0.0620) & 0.0702(0.0157) & 0.4182(0.1674) & \textbf{0.0136(0.0079)} \\
          & (10,10) & 0.0495(0.0085) & 0.0395(0.0066) & 0.0499(0.0091) & \textbf{0.0084(0.0057)} \\
    \midrule
    \multirow{2}[2]{*}{DMOP2} & (5,10) & 0.0425(0.0101) & 0.1103(0.0207) & 0.1563(0.0126) & \textbf{0.0345(0.0036)} \\
          & (10,10) & 0.0427(0.0097) & 0.0850(0.0098) & 0.4293(0.0195) & \textbf{0.0162(0.0005)} \\
    \midrule
    \multirow{2}[2]{*}{DMOP3} & (5,10) & \textbf{0.0047(0.0081)} & 0.0512(0.0101) & 0.1717(0.0804) & 0.1734(0.0858) \\
          & (10,10) & \textbf{0.0044(0.0077)} & 0.0287(0.0123) & 0.1134(0.0079) & 0.1252(0.0143) \\
    \bottomrule
    \end{tabular}%
  \label{tab:igd}%
\end{table*}%

As the experimental results show, in Table \ref{tab:igd}, the proposed RTLP-DMOA performs better than the other three algorithms in 9 out of 16 test instances for MIGD values. It clearly shows that the proposed RTLP-DMOA performs better than the compared algorithms on FDA1, FDA4, FDA5, and DMOP3 under all configurations for the MIGD values. We can find that RTLP-DMOA achieves a good performance of MIGD values for tri-objective problems. This is because the prediction method based on the transfer learning method have a strong ability to explore complicated different solution distributions. However, it performs worse than SGEA for FDA3, DMOP1, and DMOP2 under all dynamic test settings. Experimental results of MIGD values indicate that the proposed RTLP-DMOA maintains better convergence over the other three state of the art DMOAs under most test functions.

\begin{table*}[htbp]
  \centering
  \caption{MEAN AND STANDARD DEVIATION VALUES OF MS METRIC FOR DIFFERENT DYNAMIC TEST SETTINGS}
    \begin{tabular}{cccccc}
    \toprule
    Problem &{$\tau_t$,$n_t$}   & RTLP-RM-MEDA & dCOEA & PPS   & SGEA \\
    \midrule
    \multirow{2}[2]{*}{FDA1} & (5,10) & \textbf{0.9983(0.0026)} & 0.8697(0.0249) & 0.8721(0.0333) & 0.9441(0.0378) \\
          & (10,10) & \textbf{0.9985(0.0024)} & 0.8921(0.0211) & 0.9635(0.0149) & 0.9782(0.0110) \\
    \midrule
    \multirow{2}[2]{*}{FDA2} & (5,10) & \textbf{0.9988(0.0047)} & 0.8267(0.0505) & 0.9013(0.0497) & 0.9934(0.0053) \\
          & (10,10) & \textbf{0.9939(0.0036)} & 0.8672(0.0285) & 0.9356(0.0121) & 0.9930(0.0034) \\
    \midrule
    \multirow{2}[2]{*}{FDA3} & (5,10) & 0.8809(0.0035) & 0.5031(0.0427) & 0.6001(0.0404) & \textbf{0.8843(0.0711)} \\
          & (10,10) & 0.8585(0.0253) & 0.5873(0.0356) & 0.6180(0.0299) & \textbf{0.9437(0.0775)} \\
    \midrule
    \multirow{2}[2]{*}{FDA4} & (5,10) & \textbf{1.0000(0.0000)} & 0.9649(0.7774) & 0.9984(0.0008) & 0.9997(0.0001) \\
          & (10,10) & \textbf{1.0000(0.0000)} & 0.9702(0.0063) & 0.9990(0.0001) & 0.9996(0.0001) \\
    \midrule
    \multirow{2}[2]{*}{FDA5} & (5,10) & \textbf{1.0000(0.0000)} & 0.9304(0.0380) & 0.9974(0.0024) & 0.9997(0.0001) \\
          & (10,10) & \textbf{1.0000(0.0000)} & 0.9551(0.0369) & 0.9979(0.0039) & 0.9995(0.0001) \\
    \midrule
    \multirow{2}[2]{*}{DMOP1} & (5,10) & \textbf{0.9961(0.0011)} & 0.8643(0.0414) & 0.9301(0.0667) & 0.9555(0.0305) \\
          & (10,10) & 0.9823(0.0006) & 0.8881(0.0255) & 0.9782(0.0339) & \textbf{0.9849(0.0179)} \\
    \midrule
    \multirow{2}[2]{*}{DMOP2} & (5,10) & \textbf{0.9962(0.0028)} & 0.7556(0.0563) & 0.8513(0.0139) & 0.9502(0.0130) \\
          & (10,10) & \textbf{0.9980(0.0263)} & 0.8145(0.0253) & 0.9600(0.0147) & 0.9810(0.0004) \\
    \midrule
    \multirow{2}[2]{*}{DMOP3} & (5,10) & \textbf{0.9969(0.0013)} & 0.8782(0.0136) & 0.8559(0.0315) & 0.5031(0.0248) \\
          & (10,10) & \textbf{0.9991(0.0014)} & 0.9104(0.0093) & 0.8880(0.0183) & 0.5838(0.0296) \\
    \bottomrule
    \end{tabular}%
  \label{tab:ms}%
\end{table*}%

It can be clearly found from the Table \ref{tab:ms} that the proposed RTLP-DMOA obtains the best results in 13 out of 16 instances for MS values. Apart from FDA3 and DMOP1, RTLP-DMOA performs better than the compared algorithms under all configurations. It is worth noting that RTLP-DMOA achieves the maximum value of MS on tri-objective problems: FDA4 and FDA5. Nevertheless, RTLP-DMOA is a little worse than SGEA on FDA3. Overall, the diversity of solutions obtained by RTLP-DMOA are extremely better than the other three algorithms in most case.

\subsection{Discussion}

In this subsection, we perform a comparative experiment to verify whether the combination with the regression transfer learning prediction can improve performance. We compare RTLP-RM-MEDA with RM-MEDA. RM-MEDA is originally used to solve the static multi-objective problem and not applicable for DMOPs. Table \ref{tab:igd5} indicates that RTLP-RM-MEDA performs better than RM-MEDA in all test functions at $n_t=10$ and $\tau_t=5$ configuration for MIGD values. The RTLP-RM-MEDA improves the RM-MEDA for MIGD values by 22.66\%–96.39\%. Table \ref{tab:ms5} indicates that RTLP-RM-MEDA performs better than RM-MEDA in all test instances for MS values. RTLP-RM-MEDA improves the RM-MEDA for MS values by 0.08\%–39.88\%. The ablation study reveals that the designed regression transfer learning prediction can significantly improve the performance of SMOAs.


\begin{table}[htbp]
  \centering
  \caption{MEAN AND STANDARD DEVIATION VALUES OF MIGD METRIC FOR DIFFERENT PROBLEMS AT $n_t=10$ AND $\tau_t=5$}
    \begin{tabular}{ccccccccc}
    \toprule
    Problem & RM-MEDA & RTLP-RM-MEDA \\
      \midrule
    FDA1  & 0.1309(0.0287) & \textbf{0.0051(0.0013)} \\
      \midrule
    FDA2  & 0.1429(0.0333) & \textbf{0.0228(0.0046)} \\
      \midrule
    FDA3  & 0.2110(0.0285) & \textbf{0.1425(0.0066)} \\
      \midrule
    FDA4  & 0.1691(0.0140) & \textbf{0.1116(0.0092)} \\
      \midrule
    FDA5  & 0.5522(0.0160) & \textbf{0.3615(0.0027)} \\
      \midrule
    DMOP1 & 0.4187(0.0689) & \textbf{0.0469(0.0620)} \\
      \midrule
    DMOP2 & 0.0696(0.0149) & \textbf{0.0425(0.0101)} \\
      \midrule
    DMOP3 & 0.0235(0.0125) & \textbf{0.0047(0.0081)} \\
    \bottomrule
    \end{tabular}%
  \label{tab:igd5}%
\end{table}%



\begin{table}[htbp]
  \centering
  \caption{MEAN AND STANDARD DEVIATION VALUES OF MS METRIC FOR DIFFERENT PROBLEMS AT $n_t=10$ AND $\tau_t=5$}
    \begin{tabular}{ccccccccc}
    \toprule
    Problem & RM-MEDA & RTLP-RM-MEDA \\
      \midrule
    FDA1  & 0.8515(0.0365) & \textbf{0.9983(0.0026)} \\
      \midrule
    FDA2  & 0.9447(0.0098) & \textbf{0.9988(0.0047)} \\
      \midrule
    FDA3  & 0.6634(0.1033) & \textbf{0.8809(0.0035)} \\
      \midrule
    FDA4  & 0.9992(0.0002) & \textbf{1.0000(0.0000)} \\
      \midrule
    FDA5  & 0.9988(0.0002) & \textbf{1.0000(0.0000)} \\
      \midrule
    DMOP1 & 0.7121(0.0759) & \textbf{0.9961(0.0011)} \\
      \midrule
    DMOP2 & 0.9274(0.0190) & \textbf{0.9962(0.0028)} \\
      \midrule
    DMOP3 & 0.9692(0.0105) & \textbf{0.9969(0.0013)} \\
    \bottomrule
    \end{tabular}%
  \label{tab:ms5}%
\end{table}%



\section{CONCLUSION}

This paper has proposed the RTLP-DMOA in solving DMOPs. When the environment has changed, a regression hypothesis which adapts to the solution distribution for predicting objective values is deduced. Then, excellent individuals are identified according to their predicted objective values and selected as an initial population, which can improve the performance of the evolutionary process.

From experimental comparison results, the proposed RTLP-DMOA is very competitive in most test functions. In our future work, we will integrate some advanced machine learning methods into evolutionary computing to enhance the evolutionary performance of existing static multi-objective optimization algorithms and solve the real world problems\cite{Min2010Embodied}\cite{Yin2015An}.

 \section*{Acknowledgment}

This work was supported by the National Natural Science Foundation of China (Grant No.61673328) and Shenzhen Scientific Research and Development Funding Program (Grant No. JCYJ20180307123637294).

\bibliography{mybibtex11}

\end{document}